\documentclass{article}
\usepackage{graphics,graphicx,caption,float,subcaption,booktabs,xcolor,multirow,array,color,ifthen,tabu,colortbl,dblfloatfix,url,relsize,algorithm,algorithmic,amssymb,xspace,nicefrac,microtype,amsmath,amsfonts,bm}
\usepackage[sort]{natbib}
\usepackage[pagebackref=true,breaklinks=true,colorlinks=true,bookmarks=false]{hyperref}
\usepackage[capitalise,noabbrev,nameinlink]{cleveref}
\usepackage[english]{babel}
\usepackage[accepted]{icml2020}
\usepackage{enumitem}

\icmltitlerunning{One Policy to Control Them All}
\begin{document}

\twocolumn[
\icmltitle{One Policy to Control Them All:\texorpdfstring{\\}{}Shared Modular Policies for Agent-Agnostic Control}
\icmlsetsymbol{equal}{*}
\begin{icmlauthorlist}
\icmlauthor{Wenlong Huang}{ucb}
\icmlauthor{Igor Mordatch}{google}
\icmlauthor{Deepak Pathak}{cmu,fair}
\end{icmlauthorlist}
\icmlaffiliation{ucb}{UC Berkelely}
\icmlaffiliation{cmu}{CMU}
\icmlaffiliation{google}{Google}
\icmlaffiliation{fair}{Facebook AI Research}
\icmlcorrespondingauthor{Deepak Pathak}{dpathak@cs.cmu.edu}
\icmlkeywords{reinforcement learning, control, locomotion, modularity, graph network, generalization, message passing}
\vskip 0.3in
]
\printAffiliationsAndNotice{}

\begin{abstract}
\begin{hyphenrules}{nohyphenation}
Reinforcement learning is typically concerned with learning control policies tailored to a particular agent. We investigate whether there exists a single global policy that can generalize to control a wide variety of agent morphologies -- ones in which even dimensionality of state and action spaces changes. We propose to express this global policy as a collection of \textit{identical} modular neural networks, dubbed as Shared Modular Policies (SMP), that correspond to each of the agent's actuators. Every module is only responsible for controlling its corresponding actuator and receives information from only its local sensors. In addition, messages are passed between modules, propagating information between distant modules. We show that a single modular policy can successfully generate locomotion behaviors for several planar agents with different skeletal structures such as monopod hoppers, quadrupeds, bipeds, and generalize to variants not seen during training -- a process that would normally require training and manual hyperparameter tuning for each morphology. We observe that a wide variety of drastically diverse locomotion styles across morphologies as well as centralized coordination emerges via message passing between decentralized modules purely from the reinforcement learning objective.
\end{hyphenrules}
\end{abstract}

\begin{figure}[t!]
\centering
\includegraphics[width=\linewidth]{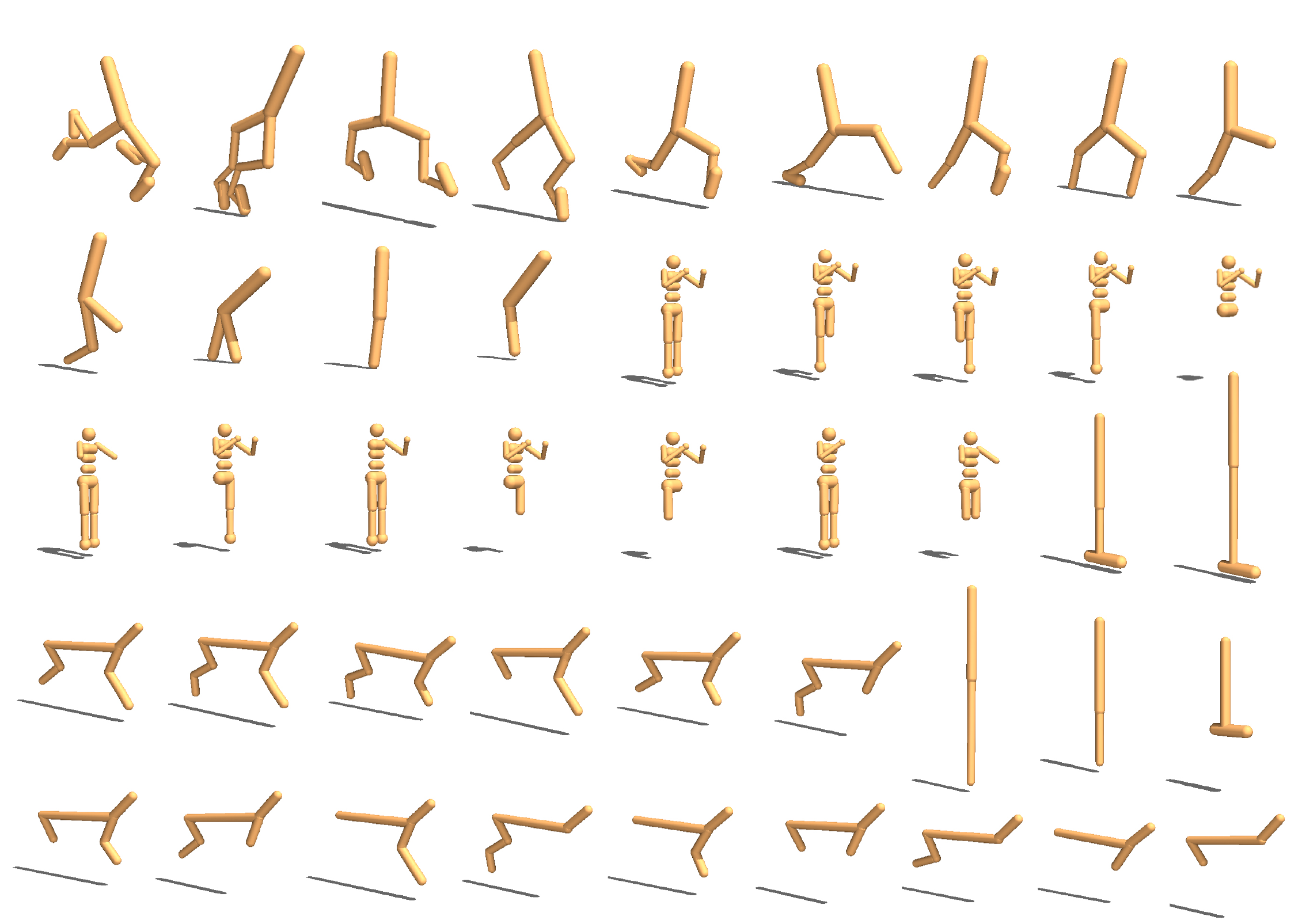}
\vspace{-.32in}
\caption{Our goal in this project is to train a single general-purpose policy controller that can perform well across many of these diverse agent morphologies.
Our key idea is to represent policy as a collection of identical and locally communicating modular neural networks shared across all limbs of all agents. Video results at~\url{https://huangwl18.github.io/modular-rl/}.}
\label{fig:teaser}
\vspace{-.05in}
\end{figure}

\section{Introduction}
\label{sec:introduction}
Deep reinforcement learning (RL) has been instrumental to successful sensorimotor control, either in simulation~\cite{heess2017emergence} or on physical robots~\cite{levine2016end}. However, state-of-the-art approaches today train a policy network from scratch that is specifically tailored to a particular robot morphology (i.e., kinematic shape) and characteristics. But if we are to ever create general, pre-trainable priors for robot control similar to those for image classification~\cite{krizhevsky2012imagenet} or natural language~\cite{devlin2018bert}, it is imperative for policies to be applicable to agents with differing morphologies.

Can a general-purpose controller be pre-trained for multiple agents by simply reducing to a multi-task RL problem? This is not easy to manifest for several reasons. Although deep RL has been proven useful in making these agents learn from scratch without any priors, their success is limited to learning a separate controller for one agent at a time with tedious hyperparameter tuning~\cite{henderson2017deep}. Moreover, unlike pre-training of vision or language models, it is difficult to contemplate as to what does pre-training a controller mean for robots because each agent may have different number of limbs, sensory inputs, motor commands, morphology, and control behaviors, as shown in Figure~\ref{fig:teaser}.

\begin{figure*}[t!]
\centering
\includegraphics[width=0.95\linewidth]{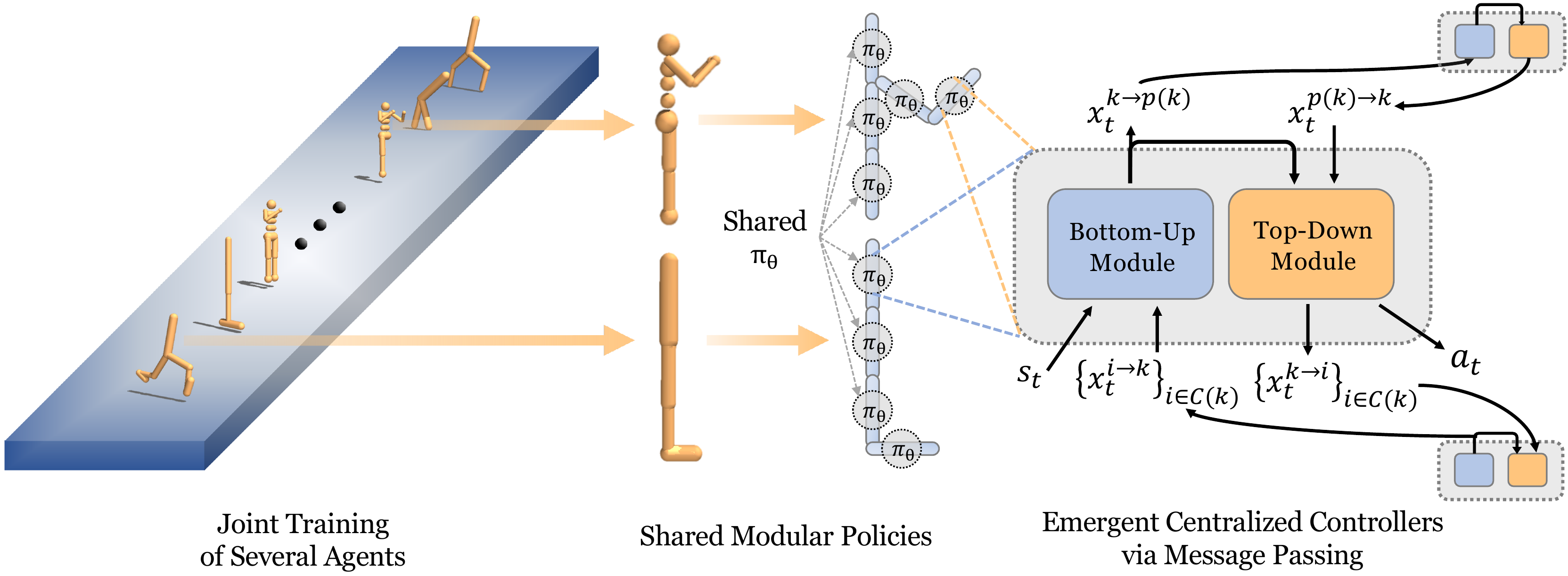}
\vspace{-.1in}
\caption{Overview of our approach: We investigate the possibility of learning general-purpose controllers by expressing agents as collections of modular components that use a shared control policy. Multiple agent controllers (left) are trained simultaneously with locally communicating modules with shared parameters (center). These modules learn to pass messages to local neighbors (right).}
\label{fig:method}
\end{figure*}

Fortunately, our natural world is abound with examples of modularity and reuse in sensorimotor systems~\cite{dickinson2000animals,holmes2006dynamics,d2015modularity}. In fact, an evidence for pre-trained controllers is seen in precocial and superprecocial animals that manage to fly or walk soon after birth, e.g. songbirds, horses, giraffe, etc.~\cite{starck1998patterns,back2013equine,fox1964phylogenetic}. At the evolutionary level, modularity allows sensorimotor design motifs to only be developed once and reused across the organism's body and propagated efficiently to its descendants. At the level of organism's lifetime, modularity offers the tools of locality and parallel processing as a means to manage complexity. Sensorimotor units only sense and act locally, e.g., a motor neuron pool may only excite a particular muscle group and only receive information from sensors physically near that muscle group~\cite{kandel2000principles}.

In contrast, current RL policies are typically centralized and holistic objects that jointly output controls for all of the agent's actuators. A centralized and holistic artificial neural network policy misses an opportunity to exploit modularity and reuse advantages both at training and execution. Can we build artificial policies that simultaneously generalize to a wide variety of agents and exploit modularity and reuse?

To answer this question, we introduce \textit{Shared Modular Policies} (SMP), a policy architecture built entirely out of a \emph{single} reusable module that is re-instantiated at each of the agent's actuators. Each module instance only perceive information from the actuator's local sensors. What makes complex coordination between modules possible is a message passing procedure where each module receives and sends learned message vectors to its neighboring actuators -- in our case neighboring limbs in the tree-structured morphology of the agent. The sensorimotor arrangement in SMP resembles a decentralized but communicating multi-agent population.  Fascinatingly, such an arrangement makes it possible to orchestrate globally coherent, coordinated behaviors, such as locomotion for complex high-dimensional agents.

SMP is trained with standard policy gradient reinforcement learning as shown in Figure~\ref{fig:method} and is able to generalize to control of variants not seen during training as we show in Section~\ref{sec:zeroshot}. This is a very hard task, as training a controller for only one of these morphologies is by itself a challenging task~\cite{islam2017reproducibility}. The idea of sharing controllers across limbs has been investigated to control self-assembling agents in~\citet{pathak18selfAssembly}. Self-assemblies allow the flexibility to evolve agent morphologies to be easier to control, and hence, learn a \textit{specialized}, yet generalizable policy. In contrast, our setup requires that the module must perform well across all morphologies to learn a \textit{unified} policy. Such a requirement ensures we learn a policy that is truly appropriate to all agents. We find message passing -- both top-down and bottom-up -- to be crucial for successful operation and show that complex communication protocols emerge that transmit information across distant limbs despite only local connectivity as in show in Section~\ref{sec:message}.

Our contributions are as follows: firstly, we present a generalizable modular policy architecture appropriate for control of arbitrary agents. Secondly, we show that resulting policies can effectively control locomotion behaviors of several planar agents simultaneously and still match the performance of corresponding oracle, i.e., state-of-the-art method trained for the individual agents. Lastly, we analyze how centralized coordination can emerge from decentralized components in the context of sensorimotor control.

\section{Learning General-Purpose Controllers}
Consider $N$ agents, each with a unique morphology. Each agent $n\in\{1\dots N\}$ contains $K_n$ different limb actuators which are connected together to constitute its overall morphological structure. Examples of such agents include half-cheetah, 2D-humanoid, etc. agents with different physical structures (Figure~\ref{fig:teaser}), all geared towards a common goal of learning to walk. The objective is to learn a single, general-purpose controller that can simultaneously be trained on all these $N$ agent morphologies via reinforcement learning and generalize to held-out morphologies in a zero-shot manner without any further finetuning. Our key insight is to employ modularity at the most fundamental level of sensorimotor learning loop, i.e., across limbs (i.e. actuators) of the agents.

\subsection{Representing Agent Morphologies as Graphs}
\label{sec:construct-graphs}
Consider a planar agent morphology that has $K$ limbs, which we represent as an undirected graph $\mathcal{G} = (\mathcal{V}, \mathcal{E})$. Each node $v_i \in \mathcal{V}$ for $i \in \{1 \dots K\}$ represents a limb of the agent, and there is an undirected edge $(v_i, v_j) \in \mathcal{E}$ if $v_i$ and $v_j$ are connected by a joint. For the brevity of method description, let's assume that the graph is connected and acyclic (i.e., a tree) with the root node be one of the limbs, although it's easy to incorporate cycles as discussed in Section~\ref{sec:centralized}. Each node/limb thus contains an actuator that controls its movement relative to its parent node/limb.

\subsection{Sensorimotor Modules}
\label{sec:no-message passing}
We develop a modular sensorimotor control policy $\pi_\theta(.)$ that is re-purposed to output the torques for each agent limb individually. The parameters $\theta$ of this module are \textit{shared} across all the limbs $k\in\{1\dots K_n\}$ of all agents $n\in\{1\dots N\}$. At each discrete timestep $t$, the policy $\pi_\theta$ for the $k^{th}$ limb of an agent $n$ receives a local sensory state of the limb $s_t^k$ as input and then outputs the torque values $a_t^k$ for the corresponding actuator controlled by this limb. Upon executing the combined action $\{a_t^k\}_{k=1}^{K_n}$ for agent $n$ at time $t$, the environment returns the next state $\{s_{t+1}^k\}_{k=1}^{K_n}$ corresponding to all individual limbs of the agent $n$ and an overall reward for the whole morphology $r_t^n(\{s_{t+1}^k\}_{k=1}^{K_n}, \{a_t^k\}_{k=1}^{K_n})$. We now discuss the joint policy optimization and how the coordination emerges within each agent as a result of modularity.

\subsection{Modular Policy Optimization}
A straightforward way to implement a modular policy architecture is to train each limb's shared policy independently to optimize the joint reward function for whole morphology. Note that each limb has its own state-space containing positions, velocity, rotation etc., see Section~\ref{sec:env-details} for details. The parameters $\theta$ of this policy $\pi_\theta$ are optimized to maximize
the joint reward via deep reinforcement learning as follows:
\begin{align}
    &\max_{\theta}\; \mathlarger{\mathbb{E}}_{\pi_\theta} \;\mathlarger{\sum}_{n=1}^N   \mathlarger{\sum}_{t=0}^{\infty}\bigg[ \gamma^t\; r_t^n\Big(\{s_{t+1}^k\}_{k=1}^{K_n}, \; \{a_t^k\}_{k=1}^{K_n}\Big) \bigg]
    \label{eq:policy} \\
    & \text{where} \; a_t^k \;=\; \pi_\theta(s_t^k) \; \text{and} \; \gamma \; \text{is the discount factor.} \nonumber
\end{align}

In case of independent modular policies, this objective is optimized such that action is produced by a policy which is conditioned on the local state of the limb. This is similar in spirit to neural module networks used for visual question answering~\cite{andreas2016neural} with the difference that our output is also modular and not just the input, i.e., each module directly outputs the limb actuation torques unlike in NLP where the output of all modules is aggregated to generate the answer.

We optimize this objective in Equation~\eqref{eq:policy} via actor-critic setup of deterministic policy gradient algorithm which is standard practice for continuous control tasks~\cite{lillicrap2015continuous}. In particular, we use the TD3 algorithm~\cite{fujimoto2018addressing}. Note that we used an off-the-shelf implementation of TD3 without much change, and our method's ability to perform across diverse morphologies stems mostly from the modularity of proposed controllers.

\section{Modular Communication}
\label{sec:message passing}
Learning a locomotion controller for walking across diverse agent morphologies, see Figure~\ref{fig:method}, is challenging for a pure reinforcement-based setup. First reason, of course, is the sheer complexity of hard joint optimization posed by this general setup. However, a bigger issue is the absence of a common gait that could perform locomotion with these agent morphologies. For instance, a bipedal walker can move efficiently with alternating walking gait while a walker with one leg (unipedal) will have to hop forward. A walker with one full leg and the other one short needs to lead with one leg and drag with the other and so on. Similar to the presence of different locomotion gaits across animals in the natural world, there exist many different locomotion gaits for our agents as different numbers of legs require different coordination with the torso and other non-locomotory limbs. Therefore, when the modular policies at each limb operate independently from other limbs, it is nearly impossible to represent different goal-directed behaviors (e.g. locomotion gaits) across different agent morphologies due to lack of ability to model coordination in absence of communication across limbs (actuators).

To facilitate limb coordination within an agent and represent different behaviors across agents, we propose to condition each limb's policy module $\pi_\theta$ on a message vector generated by its neighboring limbs in addition to conditioning on just the local state of the limb itself. Intuitively, we argue that the communication setup by these messages would enable the emergence of coherent full-body behaviors solely from identical local modules. Since our policies are fully shared, modular and communicate only with local neighbors via learned message vectors, we dub our approach as \textit{Shared Modular Policies} (SMP).

\subsection{Communication via Messages}
As discussed in Section~\ref{sec:construct-graphs}, we can represent each agent morphology as an undirected graph $\mathcal{G} = (\mathcal{V}, \mathcal{E})$. The node $k\in K_n$ in the graph corresponds to the $k^{th}$ limb and the edges denote the connectivity of limbs of agent $n$. Messages are passed along the edges between the neighboring nodes.

Let's define the torso as the root node for brevity, although any limb of the agent can act as root node of the graph and still achieve similar performance, as discussed in Section~\ref{sec:non-torso-root}. Let $p(k)$ be the unique parent of the $k^{th}$ node, $\mathcal{C}(k)$ be the set of its children nodes and $m_t^{i\rightarrow j}$ be the message from $i^{th}$ node to $j^{th}$ node. This message is a 32-dimensional learned vector generated by the limb policy.

The order in which these messages are passed governs prediction of each action $a_t^k$ in Equation~\eqref{eq:policy}. We describe three message passing schemes: bottom-up, top-down, and both-way, in which the first two are decentralized while both-way can lead to the emergence of a centralized controller.

\begin{figure*}[t!]
\centering
\includegraphics[width=\linewidth]{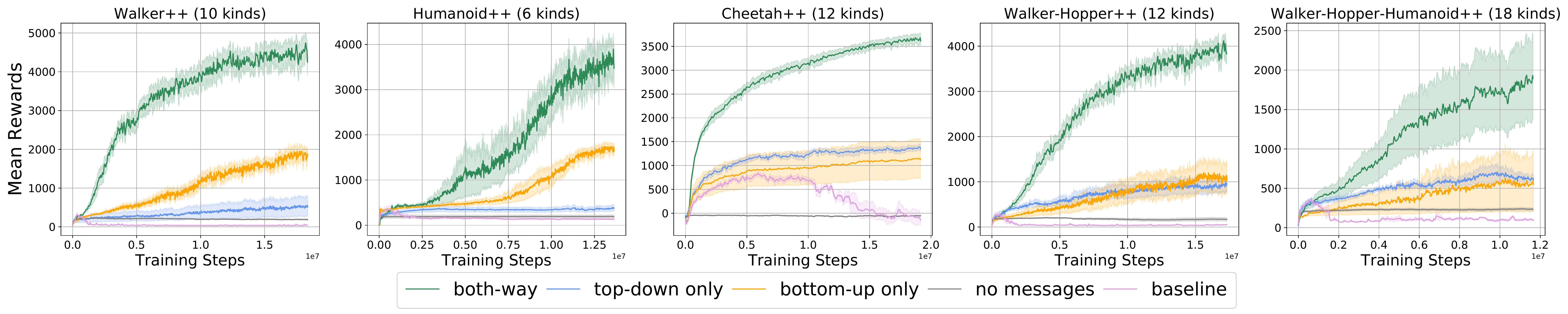}
\vspace{-.32in}
\caption{Average rewards across all agents in each of the five environments. For each environment discussed in Section~\ref{sec:env-details}, a single policy is simultaneously trained on multiple variants. Note that only 80\% of the variants are used for training and the rest 20\% are used for held-out testing. The figure shows that the multi-task baseline performs poorly on the environments containing difficult agents such as walker and humanoid, and it is also unstable in learning simpler agents like the cheetahs. In addition, the figure shows that different message passing has a pivotal impact on performance. Both-way message passing has a clear advantage in modeling diverse gaits across different agents compared to other decentralized schemes (e.g. top-down only and bottom-up only).}
\label{fig:joint}
\end{figure*}

\subsection{Decentralized Message Passing}
\label{sec:decentralized}
The communication between our modules is naturally decentralized because we have only one type of module which gets shared across all limbs. In a decentralized setup, messages can be passed either from leaf nodes to the root node, or from root node to leaf nodes, discussed as follows.

\paragraph{Bottom-Up Message Passing} Messages are passed from leaf nodes towards the root and the policy parameters $\theta$ are obtained by maximizing objective~\eqref{eq:policy} under the following constraints for action generation:
\begin{align}
    a_t^k, \; m_t^{k\rightarrow p(k)} \;=\; \pi_\theta\bigg(s_t^k,\;f\big(\{m_t^{i\rightarrow k}\}_{i\in\mathcal{C}(k)}\big) \bigg) \label{eq:bottomUp}
\end{align}
where $f(.)$ is an aggregator function that collects all the messages from child nodes and combines them into a fixed dimension output. Examples of such functions include an element-wise sum, average or max operator, etc. We discuss alternatives to aggregation in Section~\ref{sec:varyingChildren}.

\paragraph{Top-down Message Passing} Messages are passed from the root node to leaves. For simplicity, let's assume that the parent nodes passes same message output to all of its children. The policy parameters $\theta$ are trained to optimize Equation~\eqref{eq:policy} subject to following constraints on $a_t^k$:
\begin{align}
    a_t^k, \; m_t^{k\rightarrow c_k} \;=\; \pi_\theta\bigg(s_t^k,\;m_t^{p(k)\rightarrow k} \bigg) \label{eq:topDown}
\end{align}
where $m_t^{k\rightarrow c_k}$ is the message sent to all children nodes, i.e., $m_t^{k\rightarrow i} = m_t^{k\rightarrow c_k} \;\forall \;i\in\mathcal{C}(k)$. In many cases, passing a common message to all children may have issues for body-level coordination: for instance, if left and right legs have same state, then a common message won't be able to break the symmetry. To handle this, an alternative is to allow the parent to pass different messages to all its children via caching trick discussed in Section~\ref{sec:varyingChildren}.

\subsection{Both-way Message Passing: Emergence of Centralization}
\label{sec:centralized}
A purely decentralized controller should be sufficient when the diversity in morphologies is not huge and all the limb modules can converge to a \textit{similar} whole-body strategy implicitly. However, in the presence of drastically different agents like humanoid and walker, some back-and-forth communication between modules is necessary to govern a consistent full-body behavior. Although, such centralization would have to emerge and can not be encoded because our reusable design does not permit any special module which can act as a `master' node.

We allow centralization to emerge via \textbf{both-way message passing}: first from leaves to root, and then from root to leaves. The bottom-up pass generates only messages, and actions are predicted in the top-down pass. The root node can eventually learn to emerge as a centralized module that aggregates information from all other nodes and then pass on its information to others via messages. An intuitive way to understand this scheme is to draw an analogy with the central nervous system in animals where sensory neurons (upwards pass) carry information from end-effectors (leaves) to the brain (root) and then motor neurons (downward pass) carry instructions from the brain to generate output actions. To implement this, we divide our modular policy $\pi_\theta$ into two sub-policies with parameters $\theta_1$ and $\theta_2$ for the upwards pass and the downwards pass respectively. Parameters $\theta$ are trained to optimize Equation~\eqref{eq:policy} subject to following constraints:
\begin{align}
    m_t^{k\rightarrow p(k)} &\;=\; \pi_{\theta_1}\bigg(s_t^k,\;f\big(\{m_t^{i\rightarrow k}\}_{i\in\mathcal{C}(k)}\big) \bigg) \label{eq:bothways} \\
    a_t^k, \; m_t^{k\rightarrow c_k} &\;=\; \pi_{\theta_2}\bigg(m_t^{k\rightarrow p(k)},\; m_t^{p(k)\rightarrow k}\bigg) \nonumber\\
    m_t^{k\rightarrow i} &\;=\; m_t^{k\rightarrow c_k} \;\;\;\;\forall \;i\in\mathcal{C}(k) \nonumber\\
    \theta &\;=\; \{\theta_1, \;\theta_2\}  \nonumber
\end{align}
Note that the upwards pass occurs sequentially from leaf to root and only outputs a message passed to the parent. The downwards pass, in contrast, happens sequentially from root to leaf and generates the final action output and messages passed to children. The pseudo-code of the complete method is provided in the Appendix~\ref{appendix:pseudocode}.

If the morphological graph contains cycles, which however is rarely seen in the animal kingdom, the message passing can be generalized to perform multiple both-way (i.e., bottom-up and then top-down) message passing until the messages converge, as in loopy-belief-propagation~\cite{murphy1999loopy} for Bayesian networks with cycles.

\subsection{Handling Different Number of Children Nodes}
\label{sec:varyingChildren}
A parent node can have multiple children in an acyclic graph which poses a choice whether to pass the same or different messages to each child node. Section~\ref{sec:decentralized} described the scenarios when the same message is transmitted to all children nodes, which is not always optimal. For instance, when left and right legs are not symmetric and have different numbers of limbs, the torso would want to pass different latent `instruction' to each leg. In our implementation, we allow different messages via a simple caching trick where the parent node in top-down pass always outputs as many messages as the max number of child nodes across joints of all agents, i.e., $max_n K_n$. If a certain joint has fewer children, the first few distinct messages are used by each child and the remaining ones are simply ignored. A similar idea is employed in the bottom-up pass to prevent loss of information in sum or average operation over messages from different children nodes. The bottom-up policy takes $max_n K_n$ number of messages, and if the number of actual children is fewer at some node, zero vector is appended to compensate. We found that, in practice, allowing different messages between each parent-child pair in this manner works better than passing the same message to all child nodes. A generic alternative to handle different messages across child nodes is to implement the aggregator function $f(.)$ as a recurrent neural network.

\paragraph{}
The emergence of complex coordination within agent limbs by local communication between shared modules has also been explored in dynamic graph networks (DGN)~\cite{pathak18selfAssembly}. However, there are two key differences: (a) The agent shapes in our setup are static and not dynamic, thus, we do not allow the flexibility to dynamically adapt the physical morphology to make the controller learning easier. (b) Furthermore, our emphasis is on learning diverse control behaviors or gaits across these different static morphologies via different message passing mechanisms as discussed above. In contrast, in DGNs, no-message and bottom-up message passing is good enough for agents that are allowed to adjust their shape.

\begin{figure*}[t!]
\centering
\includegraphics[width=\linewidth]{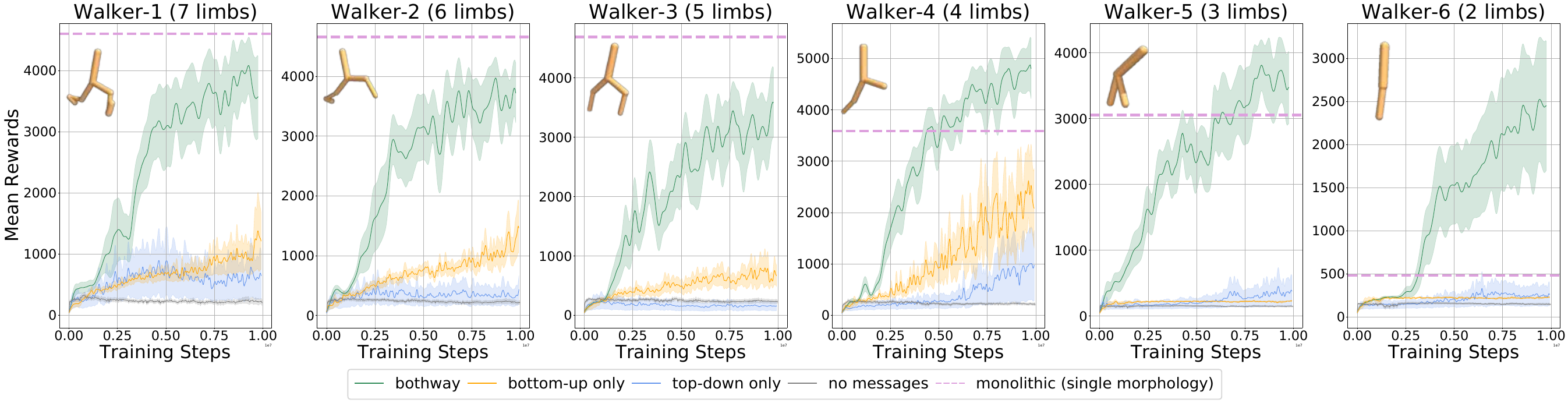}
\vspace{-.25in}
\caption{Comparisons between different message passing schemes on walker morphologies. The policy is jointly trained on ten walker variants and the figure shows a subset of six. Decentralized message passing schemes (e.g. top-down only or bottom-up only) can learn locomotion task for walkers with four to seven limbs to some extent, but fail to learn anything meaningful for three-limb and two-limb variants. In contrast, both-way message passing can model multiple gaits and demonstrates a clear advantage. Video results at~\url{https://huangwl18.github.io/modular-rl/}.}
\label{fig:walkerjoint}
\end{figure*}

\section{Experiment Setup}
We investigate our proposed general-purpose controllers on the standard Gym MuJoCo locomotion tasks. We run all environments in parallel with the shared controller across limbs. Each experiment is run with four seeds to report the mean and the standard error. The reward for each environment is calculated as the sum of instant rewards across an episode of $1,000$ time-steps.

\paragraph{Environment and Agents}
\label{sec:env-details}
We choose the following environments from Gym MuJoCo to evaluate our methods: 'Walker2D-v2', 'Humanoid-v2', 'Hopper-v2', HalfCheetah-v2'.
To facilitate the study of general-purpose locomotion principles across these agents, we modify the standard 3D humanoid to constrain it to a 2D plane similar to walker, hopper, and cheetah.

To systematically investigate the proposed method when applied to multi-task training, we construct several variants of each of the above agents, as shown in Figure~\ref{fig:method}. We create the following collections of environments using these variants: (1) 12 variants of walker [walker++], (2) 8 variants of humanoid [humanoid++], (3) 15 variants of cheetah [cheetah++], (4) all 12 variants of walker and 3 variants of hopper [walker-hopper++], and (5) all 12 variants of walker, 3 variants of hopper, and all 8 variants of humanoid [walker-hopper-humanoid++]. We keep 20\% of the variants as the held-out set and use the rest for training. Note we do not solely evaluate the methods on the hopper environment because there are only three variants possible. And we do not perform cross-category training with the cheetah environment because it uses a different integrator, making it unstable when jointly trained with other environments.

To create the variants for each agent, consider each agent as a tree with the root being the torso. We create all possible subsets (the power set) of all the nodes in the tree and keeping only those that contain the torso and form connected graphs. This can also be done by procedurally removing one leaf node at a time and enumerating all possible combinations. Note that we leave out those variants that are structurally infeasible for locomotion (e.g. humanoid without legs) in training and testing.

\paragraph{States and Actions}
The total state space of agent $n$, $\{s_{t}^k\}_{k=1}^{K_n}$, is a collection of local limb states. Each of these limb states, $s_t^k$, contain global positions, positional velocities, rotational velocities, 3D rotations, and range of movement of the limb body. We represent these 3D rotations via an exponential map representation~\cite{grassia1998practical}. The range of movement is represented as three scalar numbers $(position_t, low, high)$ normalized to $[0, 1]$, where $position_t$ is the joint position at time $t$, and $[low, high]$ is the allowed joint range.
To handle different numbers of children nodes, we implement the simple caching trick discussed in Section~\ref{sec:varyingChildren}.
Note that the torso limb has no actuator in any of these environments, so we still keep a sensorimotor module for torso for message passing but ignore its predicted torque values.

We use TD3~\cite{fujimoto2018addressing} as the underlying reinforcement learning method. The internal modules which are shared across all limbs of over 20 agents are just two 4-layered fully-connected neural networks with ReLU and tanh non-linearity, one for bottom-up message passing and the other one for top-down message passing. The dimension of message vectors is 32. Other details of training and a sanity check section that compares the Shared Modular Policies to a standard monolithic policy trained on single-agent environments can be found in the appendix.

\begin{figure*}[t!]
\centering
\includegraphics[width=\linewidth]{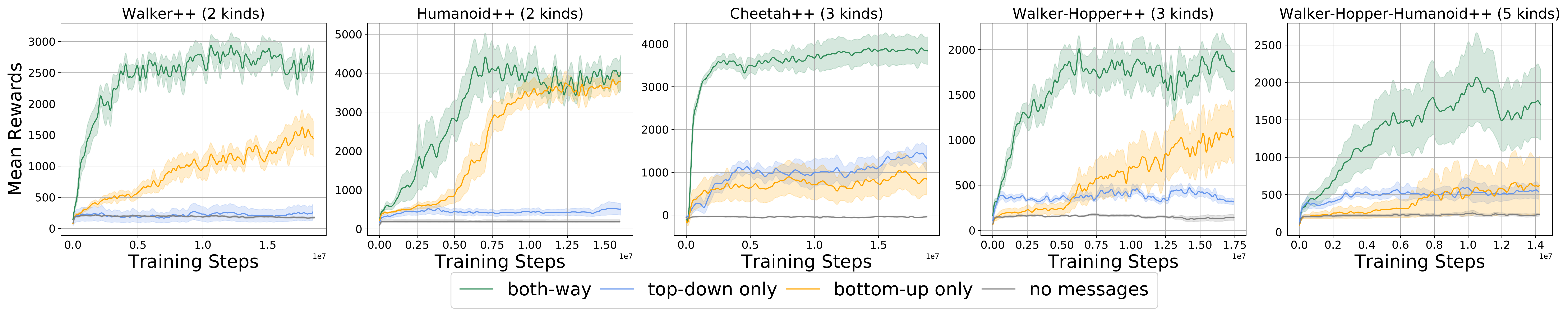}
\vspace{-.3in}
\caption{Zero-Shot Generalization: Average rewards across held-out morphologies for different message passing schemes. The policy in each plot is trained jointly on 80\% of variants from that environment and tested on the 20\% unseen variants in a zero-shot manner. We take the trained SMP policies from each timestep during training (x-axis) and test on the held-out set. SMP with both-way message passing generalizes without troubles to unseen environments, demonstrating it has learned important priors for locomotion.}
\label{fig:zeroshot}
\end{figure*}

\section{Results and Ablations}
We evaluate the effectiveness of our approach by asking three questions: Can our Shared Modular Policies (SMP) outperform the standard multi-task RL approach when simultaneously trained on many diverse agents? How do different message passing schemes compare and does centralized control emerge? Can it generalize to unseen morphologies in a zero-shot manner, a task that has been considered infeasible for the standard RL approach?
We examine these questions in three steps:
\begin{itemize}[noitemsep,topsep=0pt]
\item We first compare against the standard multi-task baseline and see how well our proposed method compares to such a monolithic policy simultaneously trained on multiple agents.
\item Next we examine the role of message passing, specifically the performance resulted from different message passing schemes.
\item Finally we test our learned modular policy on unseen agent morphologies in a zero-shot manner.
\end{itemize}
Also, we examine whether Shared Modular Policies are robust to the choice of root node while constructing the kinematic graph for message passing by choosing non-torso limbs as the root.

\subsection{Multi-Task RL Baseline}
Following the setup by~\citet{chen2018hardware}, the baseline that we compare to is a standard monolithic RL policy trained on all environments with TD3. The state space for each environment consists of the state of the agent in joint-coordinate (as in most existing methods) plus a task descriptor containing the number of limbs present and a one-hot environment ID. The policy is a four-layered fully-connected neural network. For each agent, it takes the state of the entire agent as input and outputs the continuous torque values for all the actuators. Note that the dimensions of the state space and the action space differ across different environment, so we zero-pad the states and actions to the maximum dimension across all environments.

As shown in Figure~\ref{fig:joint}, the multi-task baseline fails to perform well in any environment, possibly due to the diversity of the agents and hence the difficulty of learning a single controller for all the agents. In contrast, SMP with both-way message passing can model many different gaits across these drastically different agents.

\subsection{Role of Message Passing}
\label{sec:messages-exp}

As shown in Figure~\ref{fig:joint}, different schemes of message passing have a significant impact on the performance of the morphologies. Not only does the both-way message passing scheme outperforms the multi-task RL baseline, but it performs significantly better than the decentralized message passing schemes (e.g. top-down only and bottom-up only).

Figure~\ref{fig:joint} shows the superiority of both-way message passing in obtaining higher average rewards across a number of agents, yet it does not show in what ways both-way message passing is superior than decentralized message passing schemes. To investigate this, we plot one figure for each morphology in Walker++, where all the agent morphologies are trained with a single policy. As shown in Figure~\ref{fig:walkerjoint}, although decentralized message passing schemes seem to work in few morphologies, they fail to model different types of motion as these morphologies exhibit drastically different gaits (e.g. a two-limb walker can only hop forward). Both-way message passing, on the other hand, learns these gaits simultaneously, a task that is even infeasible by the formulation of most RL methods.

\subsection{Zero-Shot Generalization}
\label{sec:zeroshot}
There are several examples in animal kingdom where locomotion abilities are present at birth (i.e. almost `zero-shot'), for instance, foals start to walk soon after they are born~\cite{back2013equine,fox1964phylogenetic}. Similarly, our goal of learning a general-purpose controller is not limited to training morphologies but also to generalize to new ones in a zero-shot manner without any further training.

During test time, the modular policies can potentially adapt to many morphological structures, and in this section, we test the trained policy on a set of held-out agent morphologies. As shown in Figure~\ref{fig:zeroshot}, both-way message passing has a definitive advantage in generalization, achieving high rewards even in a zero-shot manner. This demonstrates that it can generalize to a wide variety of different morphologies with no fine-tuning, showing it has learned important priors for locomotion -- a key step towards learning general-purpose controllers. Please look at the success as well as the failure videos on the project website~\footnote{\url{https://huangwl18.github.io/modular-rl/}}.

\begin{figure*}[t!]
\centering
\includegraphics[width=\linewidth]{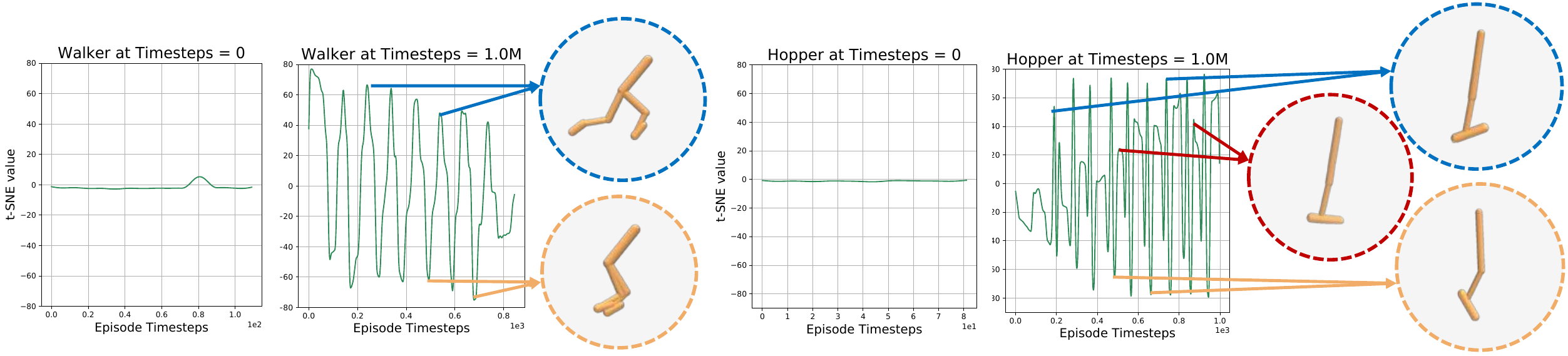}
\vspace{-.3in}
\caption{We investigate whether messages capture the alternating gait corresponding to locomotion behaviors. For both walker and hopper environment, we run t-SNE to plot the top-down root node messages against timesteps in an episode. The cyclic pattern show that the root node message not only captures such alternating gait but also plays an important role in governing the overall agent pose.}
\label{fig:tsne}
\end{figure*}

\subsection{Training with a Non-Torso Limb as Root}
\label{sec:non-torso-root}
As our method operates on a per-actuator level, it relies on the graphical representation of an agent's morphological structure, which is often a tree. Most MuJoCo environments come with such morphological structures by defining agents as an acyclic graph where torso is the root. In all the experiments in the previous sections, we simply adopt the built-in structure for each environment. However, we note that our method is agnostic to where the root is defined. To verify this, we construct another walker environment where the root is the left foot instead of the torso. We run four different seeds for the same walker morphology with this foot-root setup, the default torso-root setup, and the monolithic baseline. We report the mean and standard deviation of training rewards at 1M timesteps. Note that treating left foot as root even performs slightly better.

\vspace{0.1in}
\begin{tabular}{|c | c|}
\hline
\textbf{Method} & \textbf{Training Reward} \\
\hline
Ours (both-way) + root is left foot & $\textbf{3709.87} \pm 580.87$ \\
\hline
Ours (both-way) + root is torso & $3215.04 \pm 447.82$ \\
\hline
Monolithic Baseline & $3592.70 \pm 111.13$ \\
\hline
\end{tabular}

\section{Analysis of Message Passing}
\label{sec:message}
Illustrated by Section~\ref{sec:messages-exp}, message passing plays a crucial role for agents to orchestrate globally coherent behaviors. However, does message passing convey contextual information essential for learning general-purpose controllers or is it purely an empowering technique for modeling high-complexity tasks? We answer this question by examining the role of message passing in this section.

\vspace{-0.1in}
\paragraph{Consistency over Time}
In many of the locomotion tasks, we repeatedly observe alternating behaviors, a result of global coordination, e.g. walker moves by alternating its two legs and hopper hops by contracting and relaxing its leg. Do our learned messages capture this essence of locomotion? We investigate this question by plotting one-dimensional t-SNE~\cite{maaten2008visualizing} of the torso message, which has aggregated global information after bottom-up message passing, over the time of an episode. As shown in Figure~\ref{fig:tsne}, a clear message pattern emerges over the course of training. Furthermore, we visualize the agent across the episode time-steps and found that the agent's pose is also highly consistent with the torso message, again proving that a centralized controller can emerge from training decentralized controllers via message passing.

\section{Related Works}
Modular approaches to control that are similar to ours have been explored by robotics and virtual evolution communities. To control customizable and reconfigurable robot platforms, ~\citet{chen2018hardware,schaff2018jointly} condition the control policy on an encoding of the robot's morphology.  ~\citet{ha2017joint} avoid learning a parametric control policy altogether and instead use trajectory optimization to control the robots. When the morphology of the robot is fixed but some pre-determined parameters vary, meta-learning can be used to adapt the policy online~\cite{al2017continuous,nagabandi2018deep} or to train with variability over parameters to make the control policy insensitive to their precise value ~\cite{akkaya2019solving}. Virtual evolution similarly requires co-adaptation of the morphology and the control mechanism. Advantages of modular control have been observed in this context by ~\citet{sims1994evolving1,cheney2014unshackling,wang2019neural,pathak18selfAssembly}.

Another recent line of work exploiting modularity and reuse in deep learning are graph-structured neural networks ~\cite{scarselli2009graph} -- see ~\citep{battaglia2018relational} for a comprehensive review. Global coordination in such graph networks is either implemented via global aggregation or decentralized message passing, as in ~\cite{gilmer2017neural,zhang2019dynamic}. In deep reinforcement learning, graph structure has typically been used to efficiently encode agent's observations (i.e. world entities and interactions) as in ~\cite{sanchez2018graph,baker2019emergent}. Exceptions are works of ~\citet{wang2018nervenet,pathak18selfAssembly}, which similarly to our work exploit graph structure present in the agent's morphology.

Our message passing modules also bear resemblance to a communicating multi-agent system. Global coordination emerging from decentralized agents was observed from deep reinforcement learning agents in ~\cite{sukhbaatar2016learning,foerster2016learning,mordatch2018emergence}. Modularity has also been observed to an important component of a biological sensorimotor organization -- see ~\cite{d2015modularity} for a review. Examples of global coordination observed in this area include central pattern generators for control of rhythmic behaviors ~\cite{marder2001central} and muscle synergies ~\cite{d2003combinations}.%

\section{Conclusion}
In this work, we have presented a policy architecture built entirely out of a single reusable module -- that while acting and sensing only locally creates globally-coordinated complex movement behaviors. Such an architecture can produce locomotion for a wide variety of agents simultaneously, even those not seen during training.
Overall, we hope that our work provides the foundation for general-purpose pre-trained priors of sensorimotor control.

\section*{Acknowledgments}
We would like to thank Alyosha Efros, Yann LeCun, Jitendra Malik, Pieter Abbeel, Hang Gao and the members of BAIR community for fruitful discussions. This work was supported in part by Google faculty research award.

\bibliography{main}
\bibliographystyle{icml2020}

\clearpage
\appendix

\section{Appendix}

\subsection{Result Videos}
We show videos for all variants trained with a single policy on the project website: \url{https://huangwl18.github.io/modular-rl/}. We recommended referring to videos to observe how our single 4-layer network policy can represent different gate behaviors across different agent morphologies. One way message passing is sometimes able to learn for more than one morphology, but cannot represent multiple gates. However, both-way message passing is able to represent multiple gates due to \textit{emergence of centralization from decentralized modules}.

\subsection{Implementation and Training}
\label{sec:implementations}
We use TD3~\cite{fujimoto2018addressing} as the underlying reinforcement learning method. The internal optimizer for TD3 is Adam~\cite{kingma2014adam}. The initial positions and velocities of each agent are randomized at the beginning of each episode. For the first $10,000$ time-steps during training, actions are uniformly sampled from the action space. The policy is trained with a learning rate of $4\mathrm{e}{-4}$, a tau of $0.046$, and a exploration noise of $0.13$. All internal modules are 4-layered fully-connected neural networks with ReLU and tanh non-linearity. The dimension of message vectors are $32$. Messages are normalized before passed to the children and the parent.

For multi-morphology training, each morphology has its own  environment and an independent replay buffer of size $1e6$. The maximum size of all the replay buffers is capped at $1e7$ and whenever there are $n > 10$ environments, each environment has a replay buffer of size of $1e7 / n$. We run all environments in parrallel using vectorized environment from OpenAI Baselines~\cite{baselines}. To speed up training and inference, we also use the dynamic batching package~\cite{illia_polosukhin_2018_1299387}  when there is no dependency between modules, e.g. when the limbs are neither an ancestor nor descendent of each other.

For each single-category training (walker++, humanoid++, hopper++, and cheetah++), we use its default reward function from Gym. For multi-category training (walker-hopper++ and walker-hopper-humanoid++), we use the reward function from walker++ which consists of x-axis displacement (distance covered by agent), alive reward, and sum of squared actions (for penalizing large action values).

\begin{figure}[t!]
\centering
\includegraphics[width=0.7\linewidth]{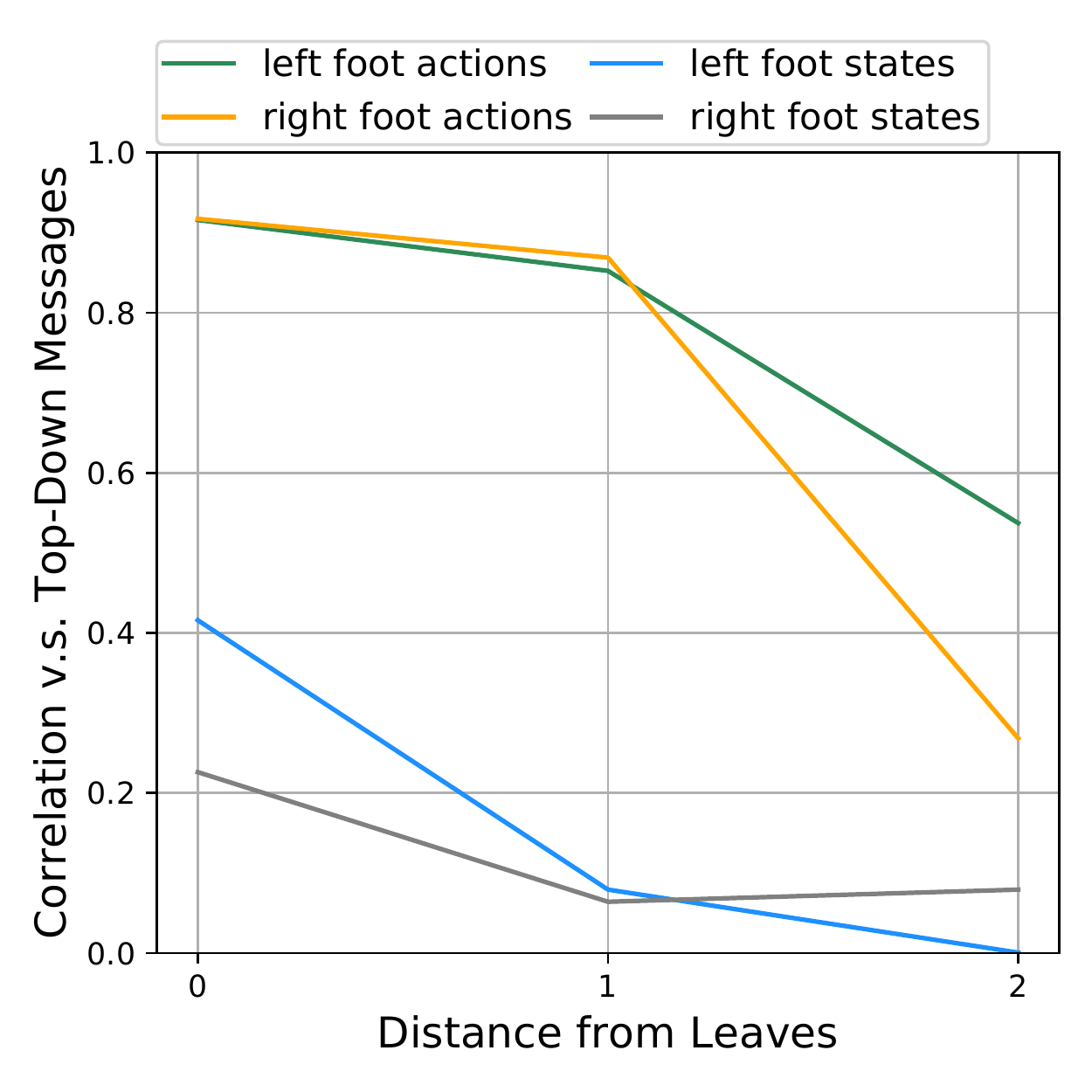}
\vspace{-.1in}
\caption{Message Range Analysis. We examine how far the messages reach and what relationship it has with the leaves' local states and actions by doing correlation analysis between leaves' local states/actions and the messages from its predecessors. The figure shows the closer a limb is to a leaf, the more its message contains the `instructions' for the leaf.}
\vspace{-0.1in}
\label{fig:correlation}
\end{figure}

\subsection{Analysis of Message Propagation Range}
In both-way message passing, messages are initially passed from the leaves to the root and then from the root back to the leaves. We here investigate whether the messages convey global information by showing the correlation between the states, actions, and the messages passed at different levels. Specifically, we test in the walker environment and we investigate how much the states and the actions are correlated with the furthest messages to the closest messages. Due to the different dimensionality, we first reduce the dimensions of the data to one by running Principle Component Analysis (PCA), and the final results are averaged over an episode. As shown in Figure~\ref{fig:correlation}, both the leaves' states and the actions are more correlated with the closer messages passed to them, demonstrating that messages are indeed conveying meaningful contextual information for locomotion.

\begin{figure*}[t!]
\centering
\includegraphics[width=\linewidth]{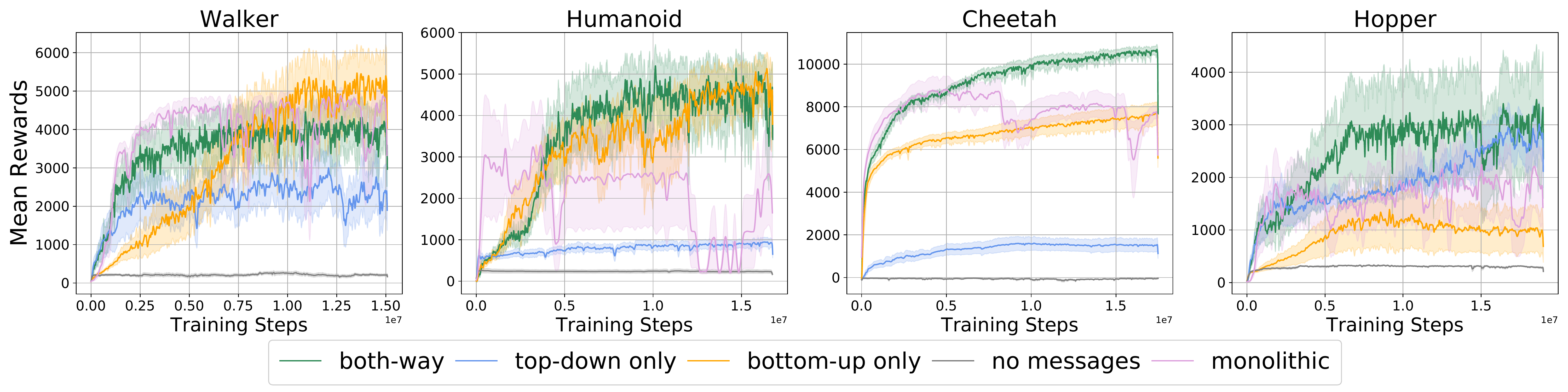}
\vspace{-.35in}
\caption{A sanity check that compares the average rewards on standard environments by the Shared Modular Policies with different message-passing schemes and the monolithic baseline. All the policies are only trained with one agent morphology at a time. The figure shows that the Shared Modular Policies with both-way message-passing can model multiple simple locomotion tasks just as well as a monolithic baseline.}
\label{fig:sanity}
\vspace{-.1in}
\end{figure*}

\subsection{Sanity Check Experiment}
We perform a sanity check on the single-agent environment to see if a modular policy can learn as well as a monolithic one based off the message passing formulation discussed in Section 3 in the main paper. It can be seen from Figure 3 of main paper that a good coordination between limbs is of the utmost importance for control because under the no message setting, all limbs act as independent agents and cannot coordinate with each other and thus cannot learn anything meaningful. This shows that a modular policy is certainly at a disadvantage when only trained and tested on a single morphology against a monolithic policy, since the latter is a much easier optimization problem and does not need to learn to generate messages just to coordinate limbs. In contrast a modular policy has to learn message passing as well as the controller to generate meaningful behaviors. Despite the challenges, we observe that, as shown in Figure~\ref{fig:sanity}, the Shared Modular Policies with both-way message-passing can achieve comparable performance to a monolithic baseline in all experiments.

\subsection{Pseudo-Code for Shared Modular Policies}
\label{appendix:pseudocode}
We provide the pseudo-code for training shared modular policies with both-way message passing. Algorithm 1 discusses the joint setup of sharing policies across motors of all agents, and Algorithm 2 discusses the end-to-end training.

\begin{algorithm}[h]
\caption{Joint Training of All Agents}
\label{alg:joint-training}
\begin{algorithmic}[1]
\STATE{\textbf{Notation Summary:}}
\STATE{$NN_{bu}$: bottom-up module parameterized by $\theta_{1}$}
\STATE{$NN_{td}$: top-down module parameterized by $\theta_{2}$}
\STATE{$s_{t}^{(e)}$: all limbs' states of environment $e$ at time $t$}
\STATE{$a_{t}^{(e)}$: all limbs' actions of environment $e$ at time $t$}
\STATE{$r_{t}^{(e)}$: rewards of environment $e$ at time $t$}
\STATE{$done^{(e)}$: whether environment $e$ is done}
\STATE{$rb^{(e)}$: replay buffer for environment $e$}
\\\hrulefill
\STATE \makebox[2em][l]{\textbf{init:}} SMP = ($NN_{bu}$, $NN_{td}$) from scratch.
\STATE \makebox[2em][l]{} empty replay buffer $rb^{(e)}$ for each environment $e$.
\WHILE{not converged}
    \STATE{\textcolor{darkgray}{// collect one episode of data for all environments}}
    \FORALL{environment $e$}
        \WHILE{$e$ is not done}
            \STATE $a_{t}^{(e)} \leftarrow $ SMP($s_{t}^{(e)}$)
            \STATE $s_{t+1}^{(e)}$, $r_{t}^{(e)}$, $done^{(e)} \leftarrow$ simulate($e$, $a_{t}^{(e)}$)
            \STATE Add ($s_{t}^{(e)}$, $s_{t+1}^{(e)}$, $a_{t}^{(e)}$, $r_{t}^{(e)}$, $done^{(e)}$) to $rb^{(e)}$
        \ENDWHILE
    \ENDFOR
    \STATE{\textcolor{darkgray}{// train SMP for each environment one by one}}
    \FORALL{environment $e$}
        \STATE SMP $\leftarrow$ trainWithTD3(SMP, $rb^{(e)}$)
    \ENDFOR
\ENDWHILE

\end{algorithmic}
\end{algorithm}

\begin{algorithm}[t]
\caption{Both-way Shared Modular Policies (SMP)}
\label{alg:smp}
\begin{algorithmic}[1]
\STATE{\textbf{Notation Summary:}}
\STATE{$NN_{bu}$: bottom-up module parameterized by $\theta_{1}$}
\STATE{$NN_{td}$: top-down module parameterized by $\theta_{2}$}
\STATE{$s_i$: local states of limb $i$}
\STATE{$a_i$: local action of limb $i$}
\STATE{$m^{i \rightarrow p(i)}$: message passed from limb $i$ to the parent of $i$}
\STATE{$m^{p \rightarrow i}$: message passed from the parent of limb $i$ to $i$}
\STATE{$\{m^{c \rightarrow i}\}_{c\in\mathcal{C}(i)}$: the set of all messages passed from the children of limb $i$ to $i$}
\STATE{$\{m^{i \rightarrow c}\}_{c\in\mathcal{C}(i)}$: the set of different messages passed from limb $i$ to each of its children}
\STATE \makebox[3.5em][l]{\textbf{Input:}} environment $e$ and all limbs' states $\{s_i\}_{i=1}^{k}$
\STATE \makebox[3.5em][l]{\textbf{Output:}} actions for all limbs of that agent $\{a_i\}_{i=1}^{k}$
\\\hrulefill

\STATE{\textcolor{darkgray}{// get agent limbs in topological order (root to leaf)}}
\STATE{nodeList $\leftarrow$ topologicalOrdering($e$)}
\STATE{\textcolor{darkgray}{// dynamically change the policy's graph structure to match that of the agent}}
\STATE{SMP $\leftarrow$ changeGraph(SMP, nodeList)}
\STATE{\textcolor{darkgray}{// bottom-up message passing (leaf to root)}}
\FOR{node $i$ in reversed(nodeList)}
    \IF{$i$ is leaf}
        \STATE{$m^{i \rightarrow p(i)} \leftarrow$} $NN_{bu}$($s_i$, $\Vec{0}$)
    \ELSE
        \STATE{$m^{i \rightarrow p(i)} \leftarrow$} $NN_{bu}$($s_i$, $\{m^{c \rightarrow i}\}_{c\in\mathcal{C}(i)}$)
    \ENDIF
\ENDFOR
\STATE{\textcolor{darkgray}{// top-down message passing (root to leaf)}}
\FOR{node $i$ in nodeList}
    \IF{$i$ is root}
        \STATE{$a_i$, $\{m^{i \rightarrow c}\}_{c\in\mathcal{C}(i)} \leftarrow$ $NN_{td}$($m^{i \rightarrow p(i)}$, $\Vec{0}$)}
    \ELSE
        \STATE{$a_i$, $\{m^{i \rightarrow c}\}_{c\in\mathcal{C}(i)} \leftarrow$ $NN_{td}$($m^{i \rightarrow p(i)}$, $m^{p(i) \rightarrow i}$)}
    \ENDIF
\ENDFOR

\STATE{\textbf{return} $\{a_i\}_{i=1}^{k}$}
\end{algorithmic}
\end{algorithm}

\end{document}